\documentclass[11pt]{article}

\usepackage[preprint]{acl}

\usepackage{times}
\usepackage{latexsym}

\usepackage[T1]{fontenc}

\usepackage[utf8]{inputenc}

\usepackage{microtype}

\usepackage{inconsolata}

\usepackage{graphicx}

\usepackage{amsmath}
\usepackage{linguex}
\usepackage{booktabs}
  \usepackage{siunitx,multirow}                                  

\title{An Existence Proof for Neural Language Models \\ That Can Explain Garden-Path Effects via Surprisal}

\author{
    \textbf{Ryo Yoshida\textsuperscript{\scriptsize $\spadesuit$}}\,
    \textbf{Shinnosuke Isono\textsuperscript{\scriptsize $\heartsuit$}}\,
    \textbf{Taiga Someya\textsuperscript{\scriptsize $\spadesuit$}}\,
    \\
    \textbf{Yohei Oseki\textsuperscript{\scriptsize $\spadesuit$,\scriptsize $\diamondsuit$}\thanks{Corresponding authors}}\,
    \textbf{Tatsuki Kuribayashi\textsuperscript{\scriptsize $\clubsuit$}\footnotemark[\value{footnote}]}
    \\
    \textsuperscript{\scriptsize $\spadesuit$}The University of Tokyo\,
    \textsuperscript{\scriptsize $\heartsuit$}NINJAL\,
    \textsuperscript{\scriptsize $\diamondsuit$}NII LLMC\,
    \textsuperscript{\scriptsize $\clubsuit$}MBZUAI
    \\
    \small{
        \texttt{\{yoshiryo0617,tsomeya,oseki\}@g.ecc.u-tokyo.ac.jp}
    }
    \\
    \small{
        \texttt{s-isono@ninjal.ac.jp}
    }
    \small{
        \texttt{tatsuki.kuribayashi@mbzuai.ac.ae}
    }
}

\begin{document}
\maketitle
\begin{abstract}
Surprisal theory hypothesizes that the difficulty of human sentence processing increases linearly with \textit{surprisal}, the negative log-probability of a word given its context. Computational psycholinguistics has tested this hypothesis using language models (LMs) as proxies for human prediction. While surprisal derived from recent neural LMs generally captures human processing difficulty on naturalistic corpora that predominantly consist of simple sentences, it severely underestimates processing difficulty on sentences that require syntactic disambiguation (\textit{garden-path effects}). This leads to the claim that the processing difficulty of such sentences cannot be reduced to surprisal, although it remains possible that neural LMs simply differ from humans in next-word prediction. In this paper, we investigate whether it is truly impossible to construct a neural LM that can explain garden-path effects via surprisal. Specifically, instead of evaluating off-the-shelf neural LMs, we fine-tune these LMs on garden-path sentences so as to better align surprisal-based reading-time estimates with actual human reading times. Our results show that fine-tuned LMs do not overfit and successfully capture human reading slowdowns on held-out garden-path items; they even improve predictive power for human reading times on naturalistic corpora and preserve their general LM capabilities. These results provide an existence proof for a neural LM that can explain both garden-path effects and naturalistic reading times via surprisal, but also raise a theoretical question: what kind of evidence can truly falsify surprisal theory?
\end{abstract}

\section{Introduction}
\label{sec:intro}
Surprisal theory~\citep{hale2001Probabilistic,levy2008Expectationbased} hypothesizes that human sentence processing involves \textit{prediction}, and that processing cost increases linearly with \textit{surprisal}, the negative log-probability of a word given its context. Computational psycholinguistics has empirically tested this hypothesis using language models (LMs), which are trained for next-word prediction on text corpora. Specifically, next-word probabilities $p_{\theta}(\text{word}\mid\text{context})$ from LMs are used as proxies for human predictability $p_{\mathrm{human}}(\text{word}\mid\text{context})$, under the principle that ``frequency affects performance''~\citep[Principle~2]{hale2001Probabilistic}. Surprisal from various LMs, including Probabilistic Context-Free Grammars (PCFGs), $n$-grams, and neural LMs, has been shown to explain human reading times and neural responses, which presumably reflect processing difficulty, providing empirical support for surprisal theory~\citep[\textit{inter alia}]{hale2001Probabilistic,smith2013Effect,frank2015ERP}.

However, while surprisal from recent neural LMs generally captures human sentence processing difficulty on naturalistic corpora that consist predominantly of simple sentences~\citep{goodkind2018Predictive,wilcox2020Predictive}, it severely underestimates processing difficulty on sentences that require syntactic disambiguation~\citep{vanschijndel2021SingleStage,huang2024Largescalea}---\textit{garden-path effects} observed in sentences like ``the horse raced past the barn fell''~\citep{bever1970cognitive}. There are two possible reasons for this failure of neural LM surprisal~\citep[][page 13]{huang2024Largescalea}. The first possible reason lies in probability estimation: the probabilities that humans and neural LMs assign to words given their context differ in some cases, resulting in the failure of neural LM surprisal to explain garden-path effects. The second possible reason lies in surprisal theory: the processing difficulty of garden-path sentences cannot be reduced to surprisal. Several recent studies argue for the second possibility~\citep{vanschijndel2021SingleStage,arehalli2022Syntactic,huang2024Largescalea,timkey2025Eye}~\citep[for a review see][Section 5]{staub2025Predictability}.

In this paper, we pursue the first possibility by investigating whether it is truly impossible to construct a neural LM that can explain garden-path effects via surprisal. While previous work has primarily evaluated off-the-shelf neural LMs, we analyze whether these LMs can be made to capture garden-path effects without overfitting or sacrificing general LM capabilities. Specifically, we fine-tune them on garden-path sentences so as to better align surprisal-based reading-time estimates with actual human reading times (Sections~\ref{sec:methods} and \ref{sec:settings}). Our results show that fine-tuned LMs do not overfit and successfully capture human reading slowdowns on held-out garden-path items; they even improve predictive power for human reading times on naturalistic corpora and preserve their general LM capabilities (Section~\ref{sec:results}). These results provide an existence proof for a neural LM that can explain both garden-path effects and naturalistic reading times via surprisal. Further analyses demonstrate that LMs fine-tuned on a single garden-path construction also capture human processing difficulty of other unseen garden-path constructions, but that the current method \textit{does not} allow the models to explain processing difficulties that are likely accounted for by memory-based theories rather than surprisal theory (Section~\ref{sec:analysis}). Finally, we discuss the theoretical implications for surprisal theory raised by our existence proof (Section~\ref{sec:discussion}).\footnote{Code for reproducing our results is available at \url{https://github.com/osekilab/RE-GPE}.}

\section{Background}
\label{sec:background}
\subsection{Surprisal Theory}
\label{subsec:Surprisal}
Surprisal theory~\citep{hale2001Probabilistic,levy2008Expectationbased} states that the processing cost of an input in human sentence processing is determined by \textit{predictability}, specifically, scaling linearly with its negative log-probability given the context:
\begin{equation*}
\text{Cost}_\mathrm{human}(w_t \mid \boldsymbol{w}_{<t}) \propto -\log p_\mathrm{human}(w_t \mid \boldsymbol{w}_{<t}).
\end{equation*}
Surprisal theory is a computational-level hypothesis in \citeposs{marr1982vision} three levels of description, providing a characterization of the goal of human sentence processing, while remaining neutral about the representations, algorithms, and implementations that realize this goal~\citep{hale2014automaton}.

Surprisal theory has several theoretical justifications. \citet{hale2001Probabilistic} showed that, given a specific grammar, surprisal equals the degree to which syntactic structures defined by that grammar are disconfirmed upon observing each new word:
\begin{equation}
\label{eq:hale}
\begin{split}
    &-\log p_\mathrm{human}(w_t \mid \boldsymbol{w}_{<t}) \\
    &= -\log \frac{\sum_{\tau \in \mathcal{T}(\boldsymbol{w}_{\leq t})} p_\mathrm{human}(\tau)}{\sum_{\tau \in \mathcal{T}(\boldsymbol{w}_{<t})} p_\mathrm{human}(\tau)},
\end{split}
\end{equation}
where $\mathcal{T}(\boldsymbol{w}_{\leq t})$ denotes the set of syntactic structures consistent with the word sequence up to position $t$. Under the assumption that ``the relation between the parser and grammar is one of strong competence''~\citep[][Principle~1]{hale2001Probabilistic}~\citep[see also][]{chomsky1965Aspects}, this degree of disconfirmation can be interpreted as processing cost---if there were processing costs distinct from those postulated in the grammar, strong competence would be violated~\citep[][Footnote 1]{hale2001Probabilistic}. This formulation requires $p_{\theta}(\tau)$ to be computed over explicit syntactic structures, such as those defined by a PCFG, to approximate $p_\mathrm{human}(w_t\mid\boldsymbol{w}_{<t})$.

\citet{levy2008Expectationbased} offered a more general interpretation, showing that without assuming a specific grammar, surprisal equals the KL divergence between posterior and prior beliefs about the latent structures $T$:
\begin{equation}
\label{eq:levy}
\begin{split}
&-\log p_\mathrm{human}(w_t \mid \boldsymbol{w}_{<t}) \\
&= D_{\mathrm{KL}}\bigl(p_\mathrm{human}(T \mid \boldsymbol{w}_{\leq t}) \,\|\, p_\mathrm{human}(T \mid \boldsymbol{w}_{<t})\bigr).
\end{split}
\end{equation}
Under the assumption that probabilities are represented as activation levels of relevant neural structures within the brain---such that larger belief updates correspond to larger physical changes~\citep[][Footnote 8]{levy2008Expectationbased}---this KL divergence can be interpreted as the cost of reallocating cognitive resources upon observing each new word. Crucially, this interpretation allows surprisal computed from string-based LMs such as $n$-grams and recent neural LMs to capture human belief updating about latent structures, as surprisal functions as a \textit{causal bottleneck} between structural representations and processing cost~\citep[][Section 2.3]{levy2008Expectationbased}.

Note that while both formulations are intended as computational-level theories, they may implicitly commit to algorithmic-level assumptions: \citeauthor{hale2001Probabilistic} interpreted his formulation as presupposing \textit{total-parallelism parsing}~\citep[][Section 3]{hale2001Probabilistic}, and \citeauthor{levy2008Expectationbased}'s interpretation is based on a specific mechanism for cognitive resource reallocation. We return to this point in Section~\ref{subsec:falsifiability}.

\subsection{Language Models as Proxies for Human Prediction}
\label{subsec:proxy}
A fundamental challenge in empirically testing surprisal theory is operationalizing human prediction $p_{\mathrm{human}}(w_t \mid \boldsymbol{w}_{<t})$, which is not directly observable.\footnote{This subsection largely follows \citet[][page 164]{levy2013Memory}.} Traditionally, researchers have employed \textit{cloze tasks}~\citep{taylor1953Cloze}, in which participants are presented with a sentence fragment and asked to predict the next word, with the proportion of participants producing each word serving as an estimate of its probability.

However, this approach suffers from a critical limitation: it cannot reliably estimate low probabilities. This is particularly problematic for surprisal theory, which assumes that processing cost scales with the \textit{logarithm} of probability, meaning that the difference between probabilities 0.0001 and 0.0099 should have the same impact as the difference between 0.01 and 0.99. Cloze tasks cannot capture such distinctions with finite sample sizes.

To address this limitation, recent work has used LMs as proxies for human prediction, assuming that corpus statistics approximate human linguistic experience and that frequency affects language processing performance~\citep[Principle~2]{hale2001Probabilistic}. Empirical results have shown that LM surprisal indeed correlates strongly with human reading times and neural responses~\citep[\textit{inter alia}]{smith2013Effect,frank2015ERP}, even outperforming cloze probability~\citep{shain2024Largescale}, providing substantial support for surprisal theory.

However, recent findings have revealed systematic discrepancies. For instance, larger and more sophisticated neural LMs exhibit \textit{worse} predictive power for human reading times despite achieving lower perplexity~\citep{oh2023Why,shain2024Largescale,kuribayashi2022Context,kuribayashi2024Psychometric}. This inverse relationship suggests that optimizing for next-word prediction does not necessarily improve alignment with human prediction. Another striking manifestation of LM-human misalignment, observed consistently across neural LMs of varying scale and architecture, is the failure to capture garden-path effects. The latter discrepancy is the main focus of this paper; we review it further in the following subsection.

\subsection{Garden-Path Effects}
\label{subsec:garden-path}
When reading sentences like Example~\ref{ex:gp-a} from left to right, humans exhibit substantially longer reading times at the word \textit{fell}~\citep[and subsequent regions reflecting spillover effects,][]{mitchell1984Evaluation}, compared to unambiguous control sentences like Example~\ref{ex:gp-b}~\citep{bever1970cognitive}:
\ex. \label{ex:gp}
\a. \label{ex:gp-a} The horse raced past the barn \textit{fell}\ldots
\b. \label{ex:gp-b} The horse \underline{that was} raced past the barn \textit{fell}\ldots

In psycholinguistics, this phenomenon is explained as follows: at the point of reading \textit{the horse raced past the barn} in Example~\ref{ex:gp-a}, a syntactic ambiguity arises between two interpretations: (i) \textit{raced} is the main verb with \textit{the horse} as its subject, and (ii) \textit{raced} forms a passive reduced relative clause, with \textit{the horse} as the modified noun. Readers prefer interpretation (i), but the appearance of \textit{fell} forces them to abandon this analysis, resulting in increased processing cost, a phenomenon known as the \textit{garden-path effect}.

Traditionally, this difficulty has been attributed to a selective reanalysis mechanism that reconstructs syntactic structures~\citep{fodor1998Reanalysisa}. However, under \citeauthor{hale2001Probabilistic}'s formulation and \citeauthor{levy2008Expectationbased}'s interpretation of surprisal theory (Equations~\ref{eq:hale} and~\ref{eq:levy}), this processing cost should be modeled as structure disconfirmation and belief updating, respectively, and thus falls within the scope of surprisal theory.

Recently, multiple studies have shown that surprisal from neural LMs consistently underestimates garden-path effects to a severe degree; for example, it predicts only approximately $1/10$ to $1/30$ of the slowdown observed in self-paced reading times~\citep{huang2024Largescalea}. This has led researchers to argue not only that next-word probabilities from neural LMs fail to serve as proxies for human predictability but also that syntactic disambiguation difficulty may be irreducible to surprisal.

\section{Methods}
\label{sec:methods}
We adopt a fine-tuning method to align surprisal-based reading-time estimates with actual human reading times~\citep{kiegeland2024ReverseEngineering}.\footnote{\url{https://github.com/samuki/reverse-engineering-the-reader}} Whereas \citeauthor{kiegeland2024ReverseEngineering} fine-tuned neural LMs on naturalistic corpora, we fine-tune neural LMs on garden-path sentences, and evaluate the resulting models on three criteria: (i) whether they generalize to held-out garden-path items without overfitting, (ii) whether they maintain predictive power for human reading times on naturalistic corpora, and (iii) whether they preserve general LM capabilities.

\paragraph{Data}
Let $D_{\mathrm{gp}}$ denote the dataset of garden-path sentences, where each data point $d \in D_{\mathrm{gp}}$ consists of a word $w_d$ and its self-paced reading time $\mathrm{RT}_d$~\citep{just1982Paradigms}. Each data point $d$ is annotated with the following attributes: sentence pair ID $s(d)$, garden-path construction $g(d) \in \{\mathrm{MVRR}, \mathrm{NPS}, \mathrm{NPZ}\}$,\footnote{Main Verb/Reduced Relative clause ambiguity, Noun Phrase/Sentential complement ambiguity, and Noun Phrase/Zero ambiguity, respectively. See Section~\ref{sec:settings} for concrete examples.} sentence ambiguity condition $c(d) \in \{\mathrm{amb}, \mathrm{unamb}\}$ (corresponding to Examples~\ref{ex:gp-a} and \ref{ex:gp-b}, respectively), position in sentence $t(d)$, and region of interest (ROI) $r(d) \in \{0, 1, 2, \mathrm{null}\}$, where $r=0$ denotes the disambiguating position (e.g., \textit{fell} in Example~\ref{ex:gp}), $r \in \{1,2\}$ denotes the two subsequent positions potentially reflecting spillover effects, and $r = \mathrm{null}$ denotes positions outside the ROI.

$D_{\mathrm{gp}}$ consists of a training set $D_{\mathrm{gp}}^{\mathrm{train}}$ and a test set $D_{\mathrm{gp}}^{\mathrm{test}}$. The training and test sets have no overlap in the verbs that induce syntactic ambiguity (e.g., \textit{raced} in Example~\ref{ex:gp}) or in words within the ROI. This ensures that the evaluation assesses whether LMs generalize to unseen data points without overfitting. We also use $D_{\mathrm{filler}}$ to denote the dataset of naturalistic filler sentences whose reading times were collected in the same experiment as $D_{\mathrm{gp}}$, and $D_{\mathrm{nat}}$ to denote a naturalistic corpus whose reading times were independently collected.

For any dataset $D$, we use $D^{(-)}$ to denote the subset excluding data points corresponding to the first two words of a sentence and sentence-final words, and $D^{(--)}$ to denote the subset further excluding data points in the ROIs ($r = 0, 1, 2$). $D^{(-)}$ can serve as the target for reading time estimation, as spillover variables are undefined for the first two words of a sentence, while sentence-final words may reflect wrap-up effects~\citep{just1980Theory}. $D^{(--)}$ is used for regression coefficient estimation. This is motivated by surprisal theory: coefficients estimated on ``ordinary'' reading times, i.e., those unaffected by syntactic disambiguation, should also account for reading times in the ROI, which are affected by disambiguation~\citep{smith2013Effect,vanschijndel2021SingleStage}.

\paragraph{Loss Function}
For each data point $d \in D_{\mathrm{gp}}^{\mathrm{train}}$, let the feature vector be $\mathbf{x}_{\theta}(d) = [\boldsymbol{\iota}_{\theta}(d)^\top, \mathbf{z}(d)^\top]^\top$, where 
\begin{equation*}
\boldsymbol{\iota}_{\theta}(d) = [-\log p_\theta(w_d^{(k)} \mid \mathrm{ctx}_d^{(k)})]_{k=0}^{2}
\end{equation*}
denotes surprisal at the current position ($k=0$) and two preceding positions ($k=1,2$) to capture spillover effects, with $w_d^{(k)}$ denoting the word at position $t(d)-k$ and $\mathrm{ctx}_d^{(k)}$ its preceding context, and $\mathbf{z}(d)$ denotes control variables such as word length.

Each batch $B \subseteq D_{\mathrm{gp}}^{\mathrm{train}}$ is sampled such that it contains equal numbers of sentence pairs from each garden-path construction. For each $B$, we estimate regression coefficients $\boldsymbol{\beta}_{\theta,B^{(--)}}$ via ridge regression:
\begin{equation*}
\begin{split}
&\boldsymbol{\beta}_{\theta,B^{(--)}} \\&= (X_{\theta,B^{(--)}}^\top X_{\theta,B^{(--)}} + \rho I)^{-1} X_{\theta,B^{(--)}}^\top \boldsymbol{\psi}_{B^{(--)}}.
\end{split}
\end{equation*}
Here, $X_{\theta,B^{(--)}}$ denotes the design matrix with rows $\mathbf{x}_{\theta}(d)^\top$ for $d \in B^{(--)}$, $\rho I$ denotes a regularization term where $I$ is the identity matrix and $\rho > 0$, and $\boldsymbol{\psi}_{B^{(--)}}$ denotes the vector of reading times $\mathrm{RT}_d$ for $d \in B^{(--)}$. We then compute the following loss:
\begin{equation}
\label{eq:loss}
\begin{split}
\mathcal{L}_B(\theta) &= \frac{1}{|B^{(-)}|} \sum_{d \in B^{(-)}} \bigl(\mathrm{RT}_d - \mathbf{x}_{\theta}(d)^\top \boldsymbol{\beta}_{\theta,B^{(--)}} \bigr)^2 \\
&\quad + \lambda \| \boldsymbol{\beta}_{\theta,B^{(--)}} - \boldsymbol{\beta}_{\theta_0,D_{\mathrm{gp}}^{\mathrm{train}(--)}} \|^2.
\end{split}
\end{equation}
The first term minimizes the squared residuals between actual and estimated reading times. The second term penalizes deviation of the regression coefficients from the initial coefficients estimated on $D_{\mathrm{gp}}^{\mathrm{train}(--)}$ using the initial LM parameters $\theta_0$.\footnote{Preliminary experiments revealed that without this term, the LM would artificially inflate estimated reading times in the ROIs by reducing surprisal outside this region to increase the regression coefficients.}

\paragraph{Evaluation}
We evaluate based on three criteria:

\subparagraph{Garden-Path Effect Alignment}
We compute regression coefficients $\boldsymbol{\beta}_{\theta,D_{\mathrm{filler}}^{(-)}}$ on $D_{\mathrm{filler}}^{(-)}$ via ridge regression as in the fine-tuning procedure. We then evaluate how well the estimated reading time difference for ambiguous versus unambiguous sentences in $D_{\mathrm{gp}}^{\mathrm{test}}$,
\begin{equation*}
\begin{split}
\Delta\widehat{\mathrm{RT}}_{g,r}(\theta) &= \frac{1}{|S_g|} \\
&\quad \times \sum_{s \in S_g} \bigl[ \mathbf{x}_{\theta}(d(s,\mathrm{amb},r))^\top \boldsymbol{\beta}_{\theta,D_{\mathrm{filler}}^{(-)}} \\
&\qquad - \mathbf{x}_{\theta}(d(s,\mathrm{unamb},r))^\top \boldsymbol{\beta}_{\theta,D_{\mathrm{filler}}^{(-)}} \bigr],
\end{split}
\end{equation*}
aligns with the actual reading time difference $\Delta \mathrm{RT}_{g,r}$~\citep{vanschijndel2021SingleStage}. Here, $S_g$ denotes the set of test pairs for garden-path construction $g$, and $d(s,c,r)$ denotes the data point corresponding to pair $s$, condition $c$, and region $r$.

\subparagraph{Impact on Naturalistic Corpora}
We evaluate the per-datapoint log-likelihood improvement of a Gaussian linear regression model including surprisal as a predictor over a baseline model with control variables only~\citep{wilcox2020Predictive}:
\begin{equation*}
\begin{split}
\Delta \mathrm{llh}(\theta) &= \frac{1}{|D_{\mathrm{nat}}^{(-)}|} \\
&\quad \times \sum_{d \in D_{\mathrm{nat}}^{(-)}} \bigl[ \log f(\mathrm{RT}_d \mid \mathbf{x}_{\theta}(d); \boldsymbol{\beta}_{\theta,D_{\mathrm{nat}}^{(-)}}) \\
&\qquad - \log f(\mathrm{RT}_d \mid \mathbf{z}(d); \boldsymbol{\beta}_{\emptyset,D_{\mathrm{nat}}^{(-)}}) \bigr],
\end{split}
\end{equation*}
where $f(\cdot \mid \cdot\,; \boldsymbol{\beta})$ denotes the probability density function of a Gaussian linear regression model with coefficients $\boldsymbol{\beta}$ and the subscript $\emptyset$ indicates the baseline regression model using control variables only. The two regression models are fitted separately on $D_{\mathrm{nat}}^{(-)}$.

\subparagraph{Language Model Capabilities}
To assess whether the fine-tuned LMs preserve general LM capabilities, we additionally evaluate perplexity on naturalistic corpora and grammatical knowledge using BLiMP~\citep{warstadt2020BLiMP}.

\begin{figure*}
    \centering
    \includegraphics[width=0.90\linewidth]{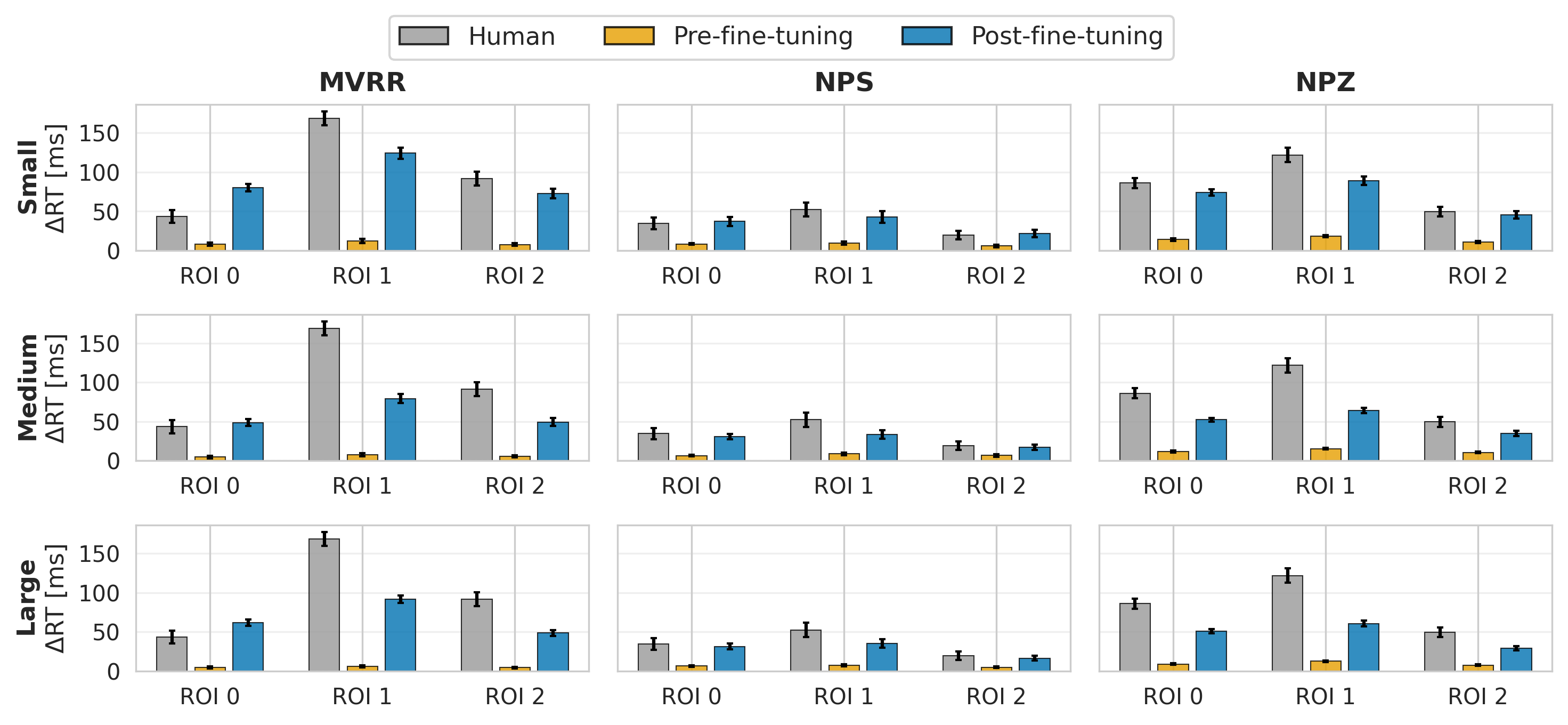}
    \caption{Garden-path effect alignment for pre- and post-fine-tuned LMs. Rows indicate model sizes and columns indicate garden-path constructions, with the x-axis representing the ROI position and the y-axis representing the reading time difference (ms) between ambiguous and unambiguous conditions. Black bars show actual human reading time differences, while orange and blue bars show estimates from pre- and post-fine-tuned LMs, respectively. Error bars represent standard errors across folds.}
    \label{fig:garden-path}
\end{figure*}

\section{Experimental Settings}
\label{sec:settings}

\paragraph{Language Models}
We use GPT-2~\citep{radford2019language} \texttt{small} (S), \texttt{medium} (M), and \texttt{large} (L) as $\theta_0$, using the Hugging Face implementation~\citep{wolf2020Transformers}.\footnote{\url{https://huggingface.co/openai-community/gpt2}} Prior work on naturalistic corpora has shown that neural LM surprisal from models around the size of GPT-2 small exhibits the best fit to human reading times~\citep{oh2023Why,shain2024Largescale}.

\paragraph{Data}
For $D_{\mathrm{gp}}$, we use the Syntactic Ambiguity Processing (SAP) dataset~\citep{huang2024Largescalea}.\footnote{\url{https://github.com/caplabnyu/sapbenchmark}} This dataset contains 24 pairs for the following three garden-path constructions.

\subparagraph{Main Verb/Reduced Relative Clause (MVRR)}
\ex. \label{ex:mvrr}
\a. \label{ex:mvrr-a} The girl fed the lamb \textit{remained} relatively calm\ldots
\b. \label{ex:mvrr-b} The girl \underline{who was} fed the lamb \textit{remained} relatively calm\ldots

This garden-path construction is similar to Example~\ref{ex:gp}: whether \textit{fed} is the main verb with \textit{the girl} as its subject, or \textit{fed} introduces a passive reduced relative clause modifying \textit{the girl}. The word \textit{remained} disambiguates ($r=0$).

\subparagraph{Noun Phrase/Sentential Complement (NPS)}
\ex. \label{ex:nps}
\a. \label{ex:nps-a} The girl found the lamb \textit{remained} relatively calm\ldots
\b. \label{ex:nps-b} The girl found \underline{that} the lamb \textit{remained} relatively calm\ldots

The ambiguity is whether \textit{the lamb} is the direct object of \textit{found} or the subject of a sentential complement. The word \textit{remained} disambiguates ($r=0$).

\subparagraph{Noun Phrase/Zero (NPZ)}
\ex. \label{ex:npz}
\a. \label{ex:npz-a} When the girl attacked the lamb \textit{remained} relatively calm\ldots
\b. \label{ex:npz-b} When the girl attacked\underline{,} the lamb \textit{remained} relatively calm\ldots

The ambiguity is whether \textit{the lamb} is the direct object of \textit{attacked} or the subject of the main clause. The word \textit{remained} disambiguates ($r=0$).

Each word is annotated with self-paced reading times from 220--440 anonymized native English speakers, with the number of participants varying by sentence. We exclude observations below 100\,ms or above 3000\,ms, following \citet{futrell2018Natural}. We use the average reading time across subjects as the representative value, following recent practices~\citep{pimentel2023Effect,oh2023Why,kuribayashi2024Psychometric}. The same preprocessing is applied to all subsequent datasets.

Since this dataset is relatively small for LM fine-tuning, we adopt leave-one-out (LOO) cross-validation. In each fold, we hold out one pair from each of the three garden-path constructions (three pairs total) as test data and construct the training set from the remaining pairs such that it satisfies the non-overlap constraint (see Section~\ref{sec:methods}). After excluding one pair containing data errors, we perform 23 folds and report the average across folds.\footnote{The training set contains an average of 1645 words across folds, comparable in size to the Provo corpus \citep{luke2018Provo} (1113 words), on which \citet{kiegeland2024ReverseEngineering} reported the effectiveness of this method.}

For $D_{\mathrm{filler}}$, we use the filler sentences from the same dataset~\citep[39 sentences extracted from the Provo corpus,][]{luke2018Provo}. For $D_{\mathrm{nat}}$, we use three corpora: Natural Stories~\citep[10 stories with syntactically diverse sentences, 485 sentences, 181 participants,][]{futrell2018Natural}, Brown~\citep[35 passages from written American English, 449 sentences, 35 participants,][]{smith2013Effect}, and UCL~\citep[3 unpublished novels from an online fiction platform, 361 sentences, 117 participants,][]{frank2013Reading}. All corpora are annotated with self-paced reading times from anonymized native English speakers.

\paragraph{Regression Variables}
The control variable vector $\mathbf{z}(d)$ includes unigram surprisal, word length, and position in sentence. To account for spillover effects, we also include values from one and two words prior for unigram surprisal and word length. Unigram surprisal is estimated using the \texttt{wordfreq} library~\citep{speer2022Rspeer}, and data points with missing frequency values are excluded from regression. For surprisal, we use the corrected sum of subword surprisals~\citep{oh2024Leading,pimentel2024How}. The details of fine-tuning hyperparameters are provided in Appendix~\ref{app:hypara}.

\section{Results}
\label{sec:results}
\paragraph{Garden-Path Effect Alignment}
Figure~\ref{fig:garden-path} shows the results for garden-path effect alignment. First, while surprisal from pre-fine-tuned LMs qualitatively captures the existence of garden-path effects ($\Delta\widehat{\mathrm{RT}}>0$), it substantially underestimates their magnitude, consistent with previous work~\citep{vanschijndel2021SingleStage,huang2024Largescalea}. For example, at ROI~1 \citep[the primary focus of analysis in prior work,][Figure 4]{huang2024Largescalea}, even GPT-2 small, which shows the most substantial effect estimates, captures only approximately 7\,\%, 19\,\%, and 15\,\% of the human reading time slowdown for MVRR, NPS, and NPZ, respectively. In contrast, surprisal from post-fine-tuned LMs shows substantially improved alignment with human reading time slowdowns on held-out test set $D_{\mathrm{gp}}^{\mathrm{test}}$. Among the LMs of different sizes, GPT-2 small achieves the best alignment, capturing approximately 73\,\%, 83\,\%, and 73\,\% of the human reading time slowdown at ROI~1 for MVRR, NPS, and NPZ, respectively. Furthermore, regarding the ordering of slowdown magnitudes across constructions, pre-fine-tuned LMs failed to match the human ordering ($\mathrm{MVRR}>\mathrm{NPZ}>\mathrm{NPS}$) at ROI~1, instead showing $\mathrm{NPZ}>\mathrm{MVRR}>\mathrm{NPS}$, whereas post-fine-tuned LMs exhibited slowdown magnitudes consistent with human data.

\begin{figure}[t]
    \centering
    \includegraphics[width=1.0\linewidth]{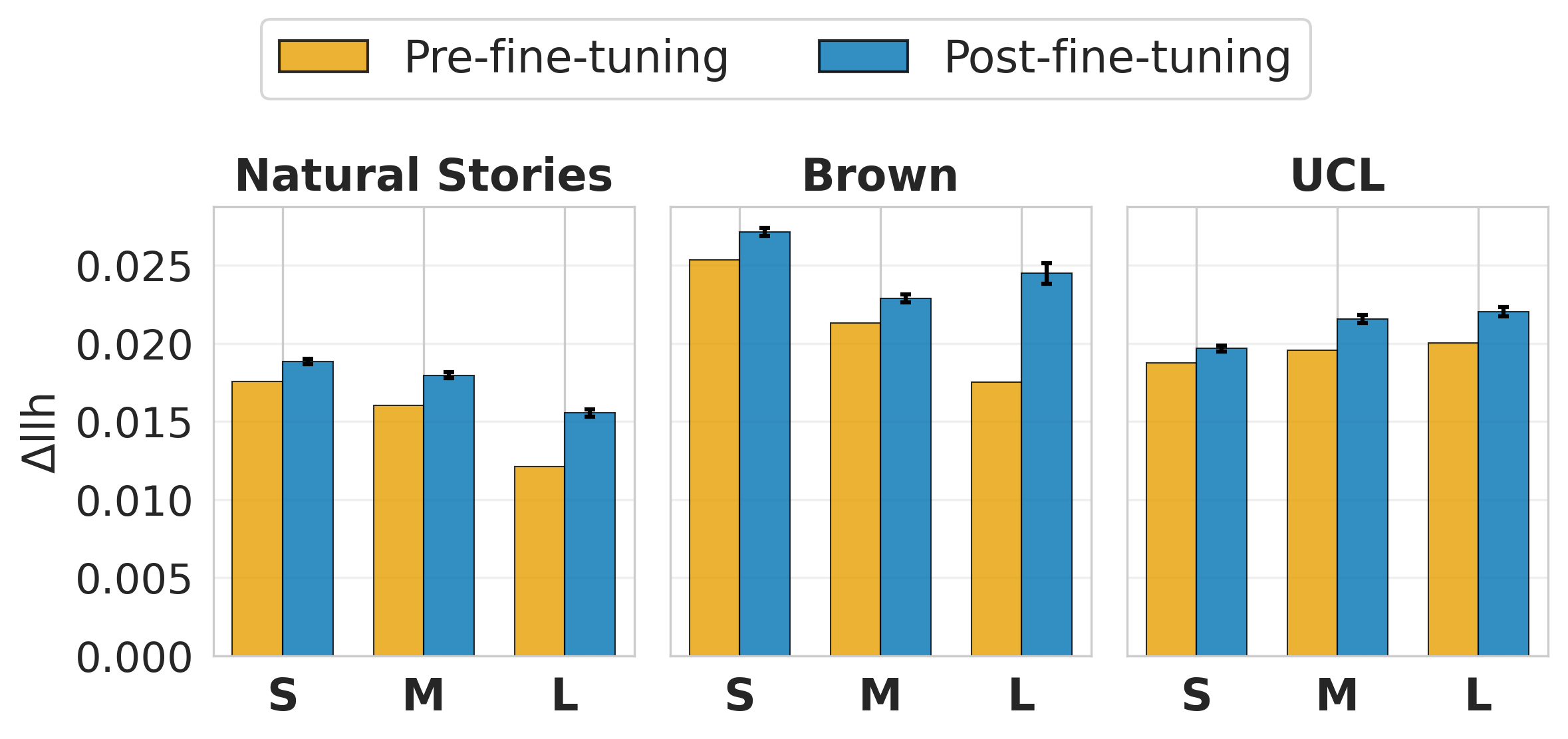}
    \caption{Impact of fine-tuning on predictive power for naturalistic corpora. Each panel corresponds to a naturalistic corpus, with the x-axis representing model size and the y-axis representing the log-likelihood improvement over the baseline regression model with control variables only.}
    \label{fig:delta_llh}
\end{figure}

\paragraph{Impact on Naturalistic Corpora}
Figure~\ref{fig:delta_llh} shows the results for the impact on naturalistic corpora. Across all corpora and all model sizes, post-fine-tuned LMs demonstrated higher predictive power for human reading times than pre-fine-tuned LMs. Interestingly, this result demonstrates that fine-tuning on garden-path sentences enhances predictive power for human reading times on naturalistic corpora that predominantly contain simple sentences.

\paragraph{Language Model Capabilities}
Table~\ref{tab:ppl_blimp_appendix} in Appendix~\ref{app:lm} shows the results for general LM capabilities. As expected, perplexity increases after fine-tuning (e.g., 53.0 $\rightarrow$ 80.0 for GPT-2 small on Natural Stories), as the training objective primarily increases surprisal. BLiMP accuracy also degrades slightly (e.g., 0.82 $\rightarrow$ 0.80 for GPT-2 small). Nevertheless, the magnitudes of degradation in both perplexity and BLiMP accuracy are comparable to those observed when fine-tuned on naturalistic reading times~\citep{kiegeland2024ReverseEngineering}, and the fine-tuned LMs remain well below the uniform perplexity baseline (50257 for all corpora) and well above chance BLiMP accuracy (0.50), indicating that general LM capabilities are largely preserved.

These results provide an existence proof for a neural LM that can explain both garden-path effects and naturalistic reading times via surprisal.

\section{Analysis}
\label{sec:analysis}
\subsection{Cross-Construction Transfer}
In Section~\ref{sec:results}, we fine-tuned neural LMs using all three garden-path constructions. In this subsection, we fine-tune them on a single garden-path construction and evaluate on all three to investigate whether the LMs learn construction-specific patterns or general mechanisms underlying garden-path effects.

\begin{figure}[t]
    \centering
    \includegraphics[width=1.0\linewidth]{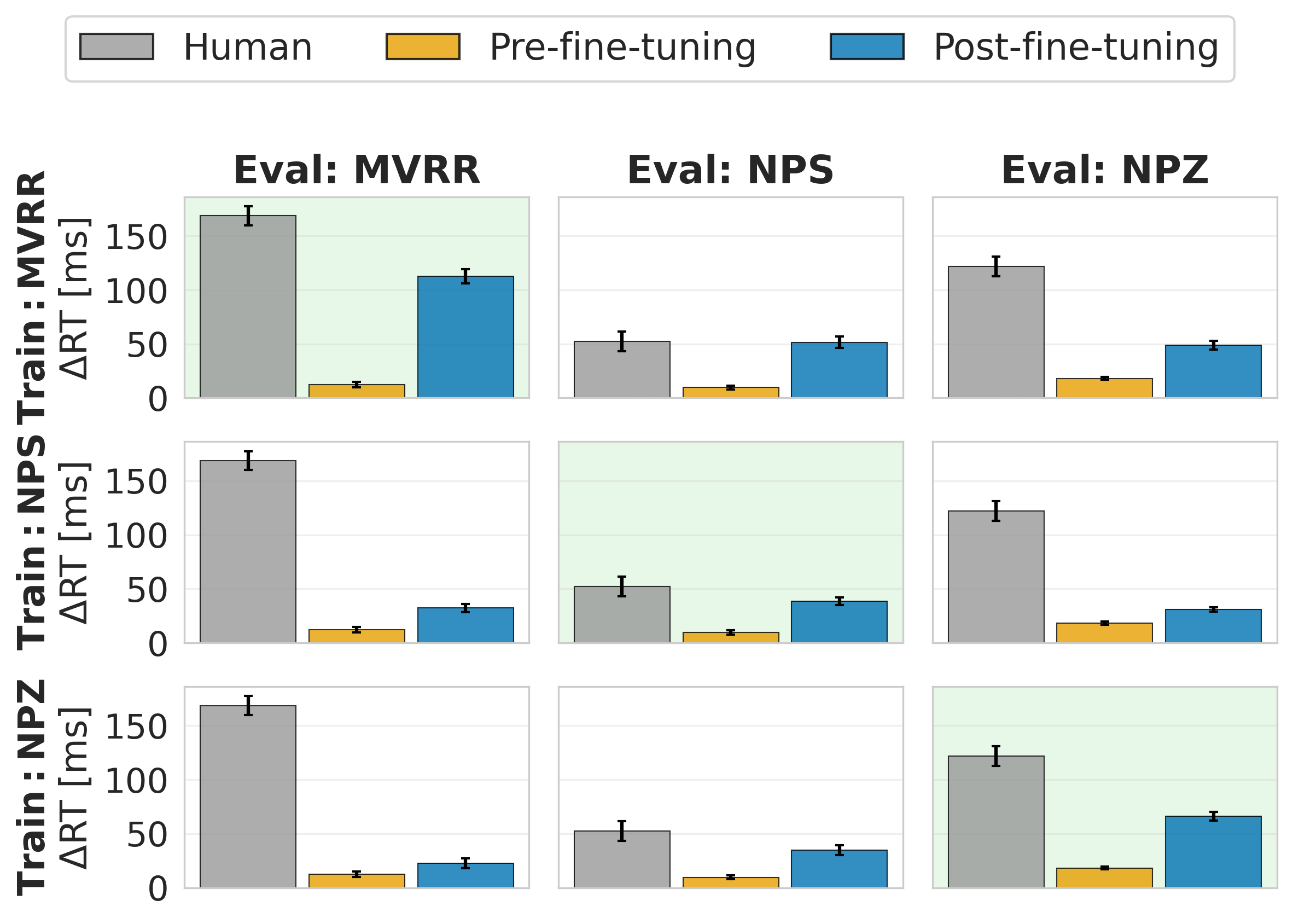}
    \caption{Cross-construction transfer for GPT-2 small at ROI~1. Rows indicate the construction used for fine-tuning, and columns indicate the construction used for evaluation, with a green background highlighting in-domain evaluation.}
    \label{fig:transfer}
\end{figure}

Figure~\ref{fig:transfer} shows the results for GPT-2 small at ROI~1. First, regarding in-domain performance, LMs fine-tuned on a single construction showed substantial improvement over the pre-fine-tuned LM, capturing 67\,\%, 73\,\%, and 54\,\% of the human slowdown for MVRR, NPS, and NPZ, respectively, though performance remained lower than when fine-tuned on all three constructions (see Figure~\ref{fig:garden-path}).

Crucially, regarding cross-construction transfer, LMs fine-tuned on one garden-path construction better captured human reading-time slowdowns on other garden-path constructions compared to the pre-fine-tuned baseline. For example, the LM fine-tuned on MVRR predicted slowdowns of 51.5\,ms (baseline: 9.6\,ms) for NPS and 48.9\,ms (baseline: 18.1\,ms) for NPZ. While the transfer is not perfect---with predictions for constructions different from the training target smaller than those from LMs fine-tuned on that construction in most cases---this result suggests that fine-tuned LMs learn general mechanisms underlying garden-path effects.\footnote{Medium and large models show broadly similar trends (Appendix~\ref{app:transfer}).}

\subsection{An Unsuccessful Example: Subject/Object Relative Clauses}
\label{subsec:src}

\begin{figure}[t]
    \centering
    \includegraphics[width=1.0\linewidth]{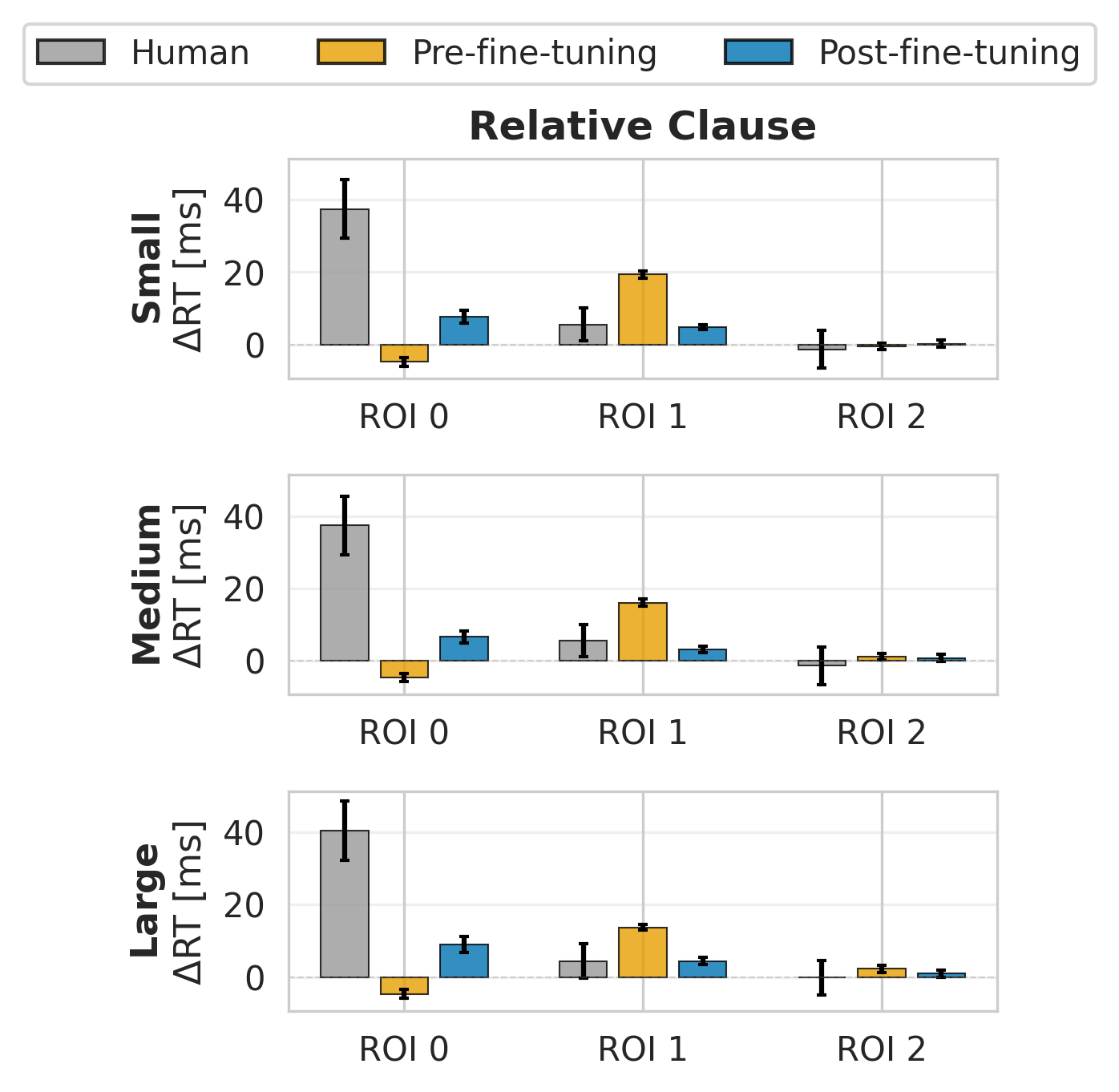}
    \caption{Subject/object relative clause asymmetry alignment for pre- and post-fine-tuned LMs}
    \label{fig:rc}
\end{figure}

One potential concern is that the current method allows neural LMs to simulate any kind of processing difficulty. If so, this would undermine the claim that there exists a neural LM that explains garden-path effects via \textit{predictability}. We address this concern by checking whether the current method also allows the models to explain processing difficulties that are unlikely to be due to predictability.

A phenomenon considered difficult to explain under surprisal theory is the processing asymmetry between English subject relative clauses (SRCs) and object relative clauses (ORCs)~\citep{levy2008Expectationbased,levy2013Memory,levy2013Surprisal}:
\ex. \label{ex:rc}
\a. \label{ex:rc-orc} The reporter that the senator \textit{attacked}\ldots
\b. \label{ex:rc-src} The reporter that \textit{attacked} the senator\ldots

Humans exhibit longer reading times at the verb position (\textit{attacked}) in ORCs like Example~\ref{ex:rc-orc} compared to SRCs like Example~\ref{ex:rc-src}. Traditional surprisal theory fails to predict this pattern: the ORC subject provides additional context that makes the verb more predictable, resulting in lower surprisal at the ORC verb than at the SRC verb. Although the increased reading time at the verb could be interpreted as a spillover effect from the unpredictable noun phrase \textit{the senator}, the dominant explanation attributes it to the increased distance to the noun phrase \textit{the reporter} that must be accessed at the point of \textit{attacked}, as posited by memory-based accounts such as Dependency Locality Theory~\citep[DLT,][]{gibson2000Dependency,grodner2005Consequences} or Category Locality Theory~\citep[CLT,][]{isono2024Category}.

In this subsection, we examine whether fine-tuning succeeds under conditions in which surprisal theory is considered inadequate, specifically for SRC/ORC asymmetry without including spillover variables, and assess its impact on predictive power for naturalistic reading times and general LM capabilities. We use 24 SRC/ORC pairs from the SAP dataset, following the original study in which the verb is designated ROI~0, the determiner ROI~1, and the noun ROI~2. The training and test sets contain completely different words at ROI~0 and ROI~2 under the same LOO setting (see Section~\ref{sec:settings}).

\begin{figure}[t]
    \centering
    \includegraphics[width=1.0\linewidth]{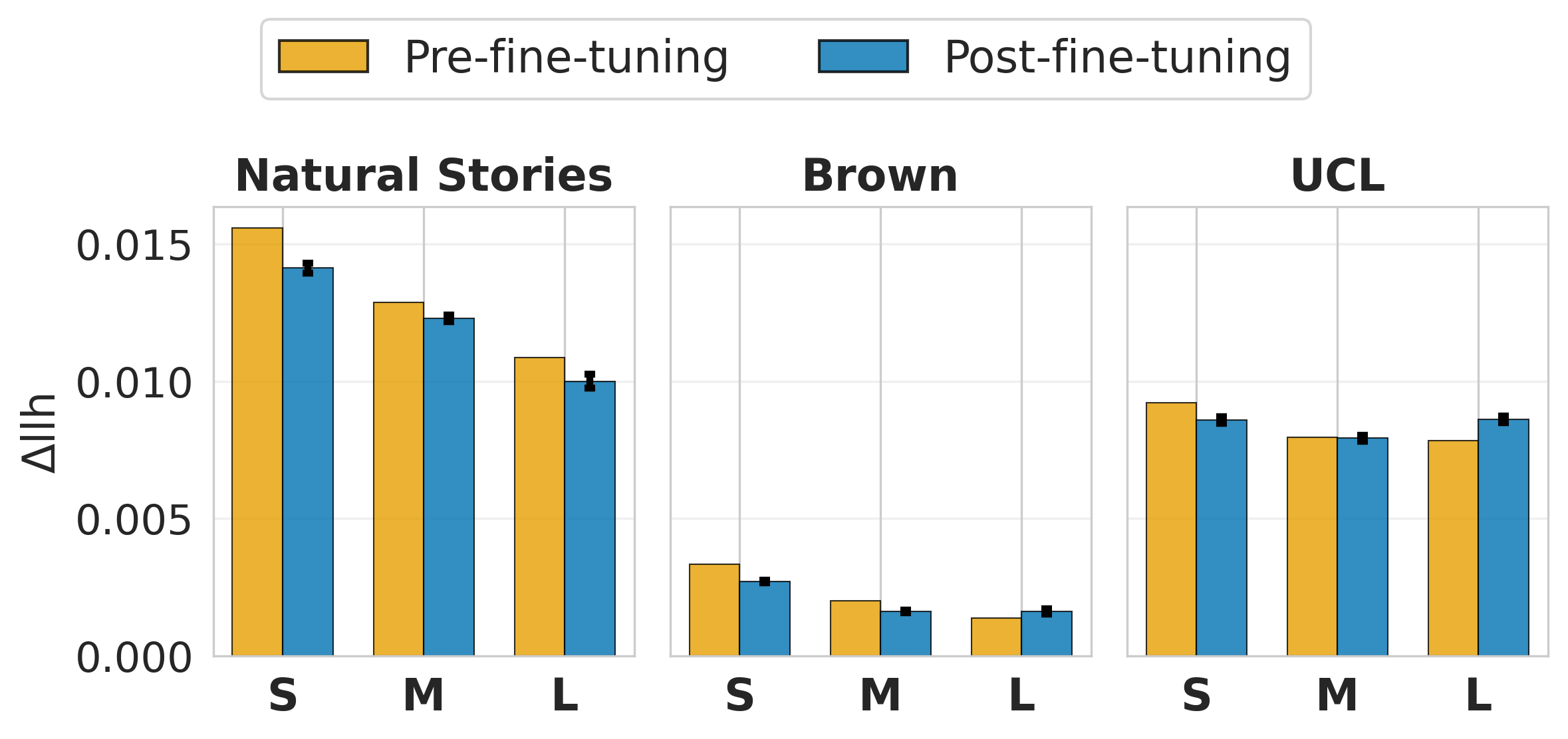}
    \caption{Impact of subject/object relative clause asymmetry fine-tuning on predictive power for naturalistic corpora}
    \label{fig:rc_delta_llh}
\end{figure}

Figure~\ref{fig:rc} shows the results.\footnote{Two of 24 folds for GPT-2 large are excluded from all evaluations due to convergence failures.} First, consistent with prior work, pre-fine-tuned LMs predict a \textit{speed-up} in reading time ($-13\,\%$ on average across model sizes) at the verb position (ROI~0), where humans show a reading time slowdown for ORCs compared to SRCs, while predicting a slowdown at the determiner position (ROI~1). Post-fine-tuned LMs learn that ORCs exhibit longer reading times than SRCs at the verb position rather than at the determiner position. However, in contrast to garden-path effects, the magnitude of the effect remains limited even for the best-performing GPT-2 large, capturing only 22\,\% of the human effect. Furthermore, as shown in Figure~\ref{fig:rc_delta_llh},\footnote{The Brown corpus, in particular, showed a substantial decrease in $\Delta$llh compared to Figure~\ref{fig:delta_llh}, as this corpus may benefit greatly from spillover predictors.} and again in contrast to garden-path effects, fine-tuning degraded predictive power for human reading times on naturalistic corpora in most conditions except for GPT-2 large on Brown and UCL. As for general LM capabilities (Table~\ref{tab:ppl_blimp_appendix} in Appendix~\ref{app:lm}), the LMs fine-tuned on SRC/ORC asymmetry show smaller perplexity degradation than the ones fine-tuned on garden-path effects (e.g., 50 $\rightarrow$ 70 for GPT-2 small on Natural Stories)---as expected, given that fine-tuning on SRC/ORC asymmetry less directly increases surprisal than fine-tuning on garden-path effects---but larger degradation in BLiMP accuracy (e.g., 0.82 $\rightarrow$ 0.78 for GPT-2 small).

These results demonstrate that, under conditions in which surprisal theory is considered inadequate, the fine-tuned LMs struggle to reproduce human reading-time differences and show degraded predictive power for naturalistic reading times.

\section{Discussion}
\label{sec:discussion}

\subsection{Implications of the Existence Proof}
\label{subsec:implications}

Several recent studies~\citep[e.g.,][]{huang2024Largescalea} have argued that given the failure of neural LM surprisal to capture garden-path effects, the processing cost of syntactic disambiguation cannot be reduced to surprisal, and consequently additional mechanisms such as reanalysis~\citep{fodor1998Reanalysisa} are necessary to account for garden-path effects. While these studies were aware that this inference is not deductively valid, noting that the space of possible LMs is unbounded~\citep[][Section 4.4]{vanschijndel2021SingleStage}, they treated the failure of current neural LMs as strong inductive evidence against surprisal theory. Our existence proof shows that this inductive argument does not hold even in practice. Note that this finding does not logically rule out the reanalysis account. Taken literally, reanalysis to a structure $\tau$ at position $t$ requires $p_{\mathrm{human}}(T=\tau|\boldsymbol{w}_{<t})=0$. This may give the impression that surprisal and reanalysis are mutually inconsistent explanations of garden-path effects. However, surprisal as a computational-level theory need not commit to the psychological reality of all the parses assigned non-zero probabilities. Under this view, surprisal and reanalysis can be parallel explanations of the garden-path effects.

More fundamentally, the unboundedness of the LM space is problematic not just for surprisal-based accounts of garden-path effects but for surprisal theory itself. Surprisal theory, at its core, merely posits the existence of \emph{some} probability distribution that log-linearly predicts human processing cost without specifying constraints that distribution must satisfy to be interpreted as human predictability. Our proof raises a troubling possibility: if the choice of training procedure is left unconstrained, the expressive power of modern neural LMs may make any such distribution empirically constructible~\citep{bowers2025studies}, rendering surprisal theory unfalsifiable in practice.\footnote{Our additional analysis on SRC/ORC asymmetry (Section~\ref{subsec:src}) suggests some empirical bounds on this flexibility, though whether this limitation persists with larger or more diverse training data remains an open question.}

\subsection{Toward a Falsifiable Surprisal Theory}
\label{subsec:falsifiability}

We argue that a promising direction is to improve the falsifiability of surprisal theory~\citep{Popper1959Logi}. We consider two directions:

\paragraph{Direction (i): Constraining the Probability Distribution}
One direction is to specify constraints that the probability distribution must satisfy, building upon the spirit of \citeposs{hale2001Probabilistic} Principles. For example, consider making Principle~2, ``frequency affects performance'', a strict constraint. One possibility is to constrain the training procedure: to what extent are interventions on the training distribution permissible? For instance, is fine-tuning with a regularization term (see Equation~\ref{eq:loss}) admissible, or must the distribution be estimated solely from a naturalistic corpus? Another is to constrain the quantity and quality of training data: given Principle~2 appeals to frequencies that humans have experienced, should training be restricted to human-scale corpora~\citep{warstadt2023Findings}?

\paragraph{Direction (ii): Requiring Psychological Reality}
A second direction is to abandon the purely computational-level stance and to require the psychological reality of the parse distribution posited in Equations~\ref{eq:hale} and~\ref{eq:levy}, as \citet{hale2001Probabilistic} and \citet{levy2008Expectationbased} appear to have implicitly done (Section~\ref{subsec:Surprisal}).\footnote{Note that this does not require always maintaining full parallelism over all theoretically possible structures. As \citet[][Section 2.4]{levy2008Expectationbased} argues, it suffices to commit to parallelism over analyses that are realistically competing in a given context---such as the main verb and reduced relative analyses in garden-path sentences.} Under this view, the neural realization of each structure and its probability, posited in the right-hand sides of these equations, becomes an empirical question, thereby grounding the falsifiability of surprisal theory in implementational evidence~\citep[see][for a parallel argument regarding the falsifiability of the \textit{Bayesian brain hypothesis},~\citealt{friston2010Freeenergy}]{mangalam2025Myth}. Note that under this direction, the reanalysis account and surprisal theory become mutually incompatible: if competing structures are mentally represented in parallel, there is no role for a selective reanalysis mechanism that constructs such structures only post-hoc.

\section{Conclusion}
\label{sec:conclusion}
In this paper, we provide an existence proof for a neural LM that can explain both garden-path effects and naturalistic reading times via surprisal, while highlighting that surprisal theory may be too flexible to falsify. We propose two directions to make surprisal theory falsifiable: by imposing strict constraints on the probability distribution, or by requiring the psychological reality of the posited parse distribution.

\section*{Limitations}
\label{sec:limitations}
This study evaluates LMs using leave-one-out cross-validation on a relatively small dataset containing three garden-path constructions. While this dataset represents the largest collection of garden-path sentences with human reading time annotations available to our knowledge, future work should validate our findings on larger-scale data. Additionally, extending this investigation to other languages and garden-path constructions would be valuable for assessing the cross-linguistic generalizability of our findings.

Our study focuses on demonstrating the existence of a neural LM that can explain both garden-path effects and naturalistic reading times via surprisal, but does not investigate the internal mechanisms that change as a result of fine-tuning. While examining such internal mechanisms falls outside the scope of our current research question, future investigations into how fine-tuning modifies internal mechanisms could provide valuable insights.

\section*{Acknowledgements}
We thank Masaki Kumakawa, Ryo Ueda, Shunsuke Kando, and Taiga Ishii for valuable discussions on this paper. This work used a generative AI tool (Claude) for language editing and coding assistance. This work was supported by JSPS KAKENHI Grant Number JP24H00087, Grant-in-Aid for JSPS Fellows JP24KJ0800, JST BOOST Grant Number JPMJBY24B2, JST CREST Grant Number JPMJCR2565, JST PRESTO Grant Number JPMJPR21C2, JST ACT-X Grant Number JPMJAX25CS, and JST SPRING Grant Number JPMJSP2108.

\bibliography{Lab}

\newpage
\appendix

\section{Hyperparameters}
\label{app:hypara}
\begin{table*}
    \centering
    \begin{tabular}{ll}
        \toprule
         Optimizer & AdamW~\citep{loshchilov2019Decoupled} \\
         LR scheduler & Cosine annealing with warm restarts \\
                      & \citep{loshchilov2017SGDR}\\
         Batch size & 66/44 \\
         Training steps & 500 \\
         Warm-up steps & 3 \\
         Max learning rate & $5.25\times10^{-5}$/$3.5\times10^{-5}$ \\
         Min learning rate & $7.8\times10^{-8}$/$5.2\times10^{-8}$ \\
         Decrease rate of max LR & 0.01\\
         Weight of loss regularization $\lambda$ & 100 \\
         Weight of ridge regularization $\rho$ & $1.0\times10^{-5}$ \\
        \bottomrule
    \end{tabular}
    \caption{Hyperparameters used for fine-tuning. For parameters with two values separated by a slash, the first value corresponds to training with all three garden-path constructions and the second to training with a single construction.}
    \label{tab:hypara}
\end{table*}

Fine-tuning hyperparameters are shown in Table~\ref{tab:hypara}. The total computational cost for all experiments was approximately 40\,GPU hours on an NVIDIA RTX 6000 Ada (48\,GB).

\section{Language Model Capabilities}
\label{app:lm}

\begin{table*}[t]
\centering
\small
\sisetup{separate-uncertainty=true, table-align-uncertainty=true}
\begin{tabular}{@{}ll
  S[table-format=5.2(2.2)]
  S[table-format=5.2(2.2)]
  S[table-format=5.2(2.2)]
  @{\hspace{1.5em}}
  S[table-format=1.3(1.3)]@{}}
\toprule
 & & \multicolumn{3}{c}{Perplexity ($\downarrow$)} & {Accuracy ($\uparrow$)} \\
\cmidrule(lr){3-5} \cmidrule(l){6-6}
Size & Condition & {Natural Stories} & {Brown} & {UCL} & {BLiMP} \\
\midrule
\multirow{3}{*}{Small}
  & Pre        & 53.04 & 78.44 & 60.95 & 0.821 \\
  & Post (GP)  & 80.08 \pm 0.93 & 130.17 \pm 1.62 & 100.34 \pm 1.62 & 0.803 \pm 0.002 \\
  & Post (RC)  & 70.10 \pm 0.47 & 107.26 \pm 0.85 & 80.29 \pm 1.00 & 0.787 \pm 0.002 \\
\addlinespace[2pt]
\multirow{3}{*}{Medium}
  & Pre        & 43.82 & 64.94 & 50.63 & 0.827 \\
  & Post (GP)  & 58.83 \pm 0.56 & 93.14 \pm 0.97 & 73.48 \pm 1.18 & 0.822 \pm 0.001 \\
  & Post (RC)  & 55.47 \pm 0.61 & 84.88 \pm 0.98 & 65.81 \pm 1.09 & 0.788 \pm 0.002 \\
\addlinespace[2pt]
\multirow{3}{*}{Large}
  & Pre        & 39.17 & 59.52 & 48.42 & 0.836 \\
  & Post (GP)  & 76.72 \pm 14.25 & 120.95 \pm 24.41 & 81.85 \pm 10.52 & 0.807 \pm 0.014 \\
  & Post (RC)  & 62.20 \pm 1.65 & 92.20 \pm 2.16 & 79.10 \pm 2.70 & 0.772 \pm 0.005 \\
\midrule
\multicolumn{2}{@{}l}{Uniform baseline}
  & \multicolumn{1}{c}{50257}
  & \multicolumn{1}{c}{50257}
  & \multicolumn{1}{c}{50257}
  & \multicolumn{1}{c}{0.500} \\
\bottomrule
\end{tabular}
\caption{Perplexity (lower is better) on three naturalistic corpora and BLiMP overall accuracy (higher is better) for pre- and post-fine-tuned LMs (GP: garden-path effects, RC: SRC/ORC asymmetry).  Values for post-fine-tuned LMs are means~$\pm$~standard errors across folds. The uniform baseline assumes a uniform distribution over the vocabulary.}
\label{tab:ppl_blimp_appendix}
\end{table*}

Table~\ref{tab:ppl_blimp_appendix} shows perplexity and BLiMP results for pre- and post-fine-tuned LMs.

\section{Full Results of Cross-Construction Transfer}
\label{app:transfer}
\begin{figure*}
    \centering
    \includegraphics[width=0.90\linewidth]{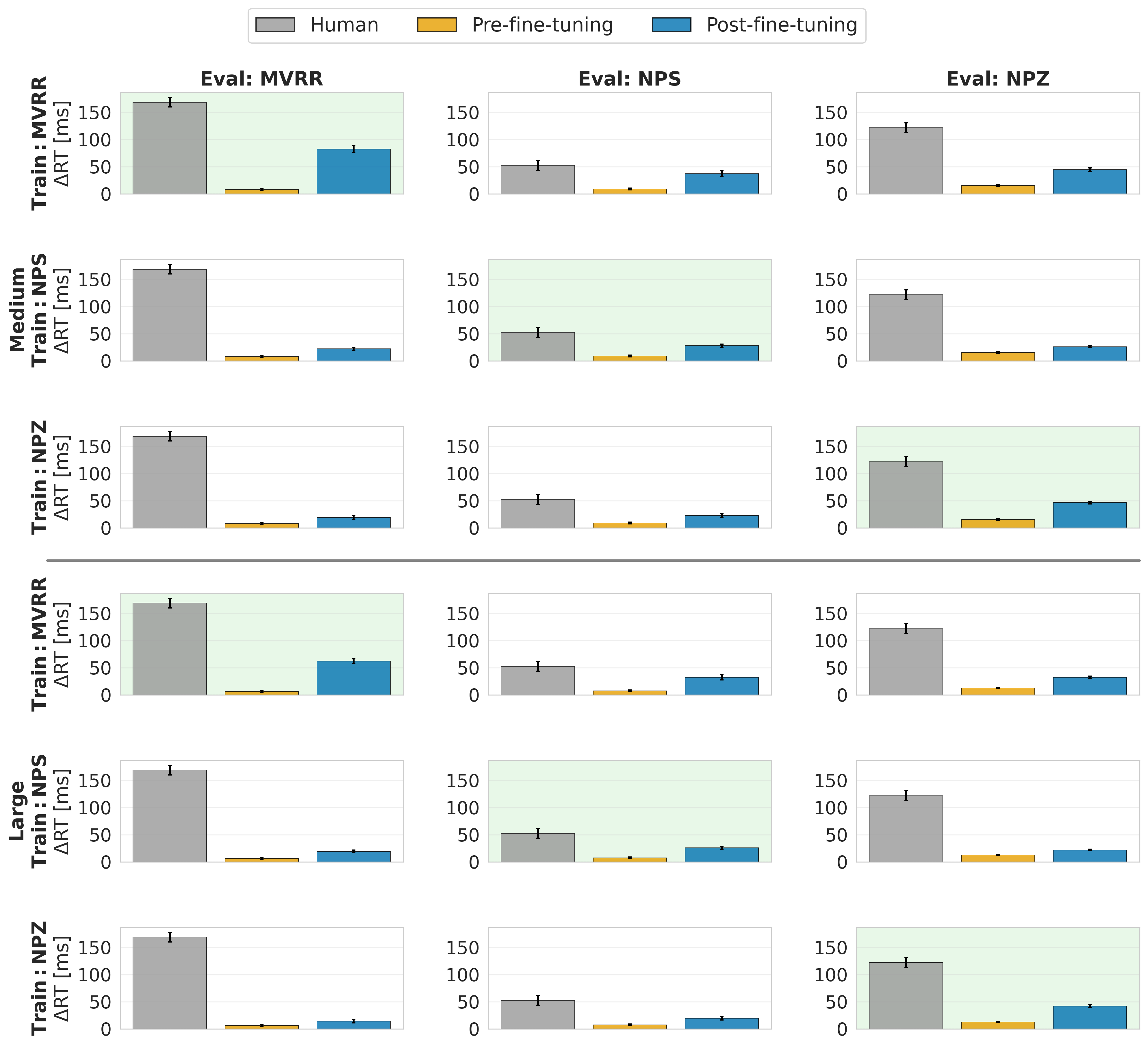}
    \caption{Cross-construction transfer results for GPT-2 medium and large at ROI~1. Within each panel, rows indicate the construction used for fine-tuning and columns indicate the construction used for evaluation, with a green background highlighting in-domain evaluation. The y-axis shows estimated reading time differences (ms) between ambiguous and unambiguous conditions.}
    \label{fig:transfer_all}
\end{figure*}
Figure~\ref{fig:transfer_all} shows cross-construction transfer results for GPT-2 medium and large at ROI~1.

\section{Licenses}
\label{app:licenses}

\begin{table*}[]
    \centering
    \begin{tabular}{ll}
        \toprule
        Dataset/Tool & License \\
        \midrule
        \multicolumn{2}{l}{\textit{Datasets}} \\
        \quad SAP dataset~\citep{huang2024Largescalea} & MIT License \\
        \quad Natural Stories corpus~\citep{futrell2018Natural} & CC BY-NC-SA 4.0 \\
        \quad Brown corpus~\citep{smith2013Effect} & CC BY 3.0 \\
        \quad UCL corpus~\citep{frank2013Reading} & CC BY 3.0 \\
        \midrule
        \multicolumn{2}{l}{\textit{Tools}} \\
        \quad \texttt{transformers}~\citep{wolf2020Transformers} & Apache 2.0 \\
        \quad \texttt{wordfreq}~\citep{speer2022Rspeer} & Apache 2.0 \\
        \bottomrule
    \end{tabular}
    \caption{Licenses of datasets and tools used in this paper}
    \label{tab:license}
\end{table*}
Table~\ref{tab:license} summarizes the licenses of the datasets and tools employed in this paper. All datasets and tools were used in accordance with their respective license terms.

\end{document}